\documentclass[sigconf]{acmart}

\usepackage{enumitem} 
\usepackage{subcaption} 
\usepackage{xspace} 

\usepackage{threeparttable} 

\usepackage{multirow}
\usepackage{makecell}

\usepackage{algorithm}
\usepackage{algpseudocode}

\usepackage[most]{tcolorbox}

\newcommand{\paratitle}[1]{\vspace{1ex}\noindent{\bf #1}}

\newcommand{\ie}{\emph{i.e.,}\xspace}

\AtBeginDocument{%
  }

\copyrightyear{2026}
\acmYear{2026}
\setcopyright{cc}
\setcctype{by}
\acmConference[KDD '26]{Proceedings of the 32nd ACM SIGKDD Conference on Knowledge Discovery and Data Mining V.2}{August 09--13, 2026}{Jeju Island, Republic of Korea}
\acmBooktitle{Proceedings of the 32nd ACM SIGKDD Conference on Knowledge Discovery and Data Mining V.2 (KDD '26), August 09--13, 2026, Jeju Island, Republic of Korea}
\acmDOI{10.1145/3770855.3817911}
\acmISBN{979-8-4007-2259-2/2026/08}

\begin{document}

\newcommand{\framework}{DyPAM\xspace}

\newcommand{\codelink}{https://github.com/Beihang-BIGSCity/DyPAM\xspace}

\title{Dynamic Positional Attention Modulation for Parameter-Efficient Fine-Tuning of Large Language Models}

\author{Dayan Pan}
\affiliation{
  \institution{School of Computer Science and Engineering,}
  \institution{MOE Engineering Research Center of Advanced Computer Application Technology,\\ Beihang University}
  \city{Beijing}
  \country{China}
}
\email{dayan@buaa.edu.cn}

\author{Jingyuan Wang}
\authornote{Corresponding authors.}
\affiliation{
  \institution{School of Computer Science and Engineering,}
  \institution{School of Economics and Management,\\ Beihang University}
  \city{Beijing}
  \country{China}
}
\email{jywang@buaa.edu.cn}

\author{Xie Yu}
\affiliation{
  \institution{School of Computer Science and Engineering,}
  \institution{MOE Engineering Research Center of Advanced Computer Application Technology, \\ Beihang University}
  \city{Beijing}
  \country{China}
}
\email{yuxie_scse@buaa.edu.cn}

\begin{abstract}
Parameter-efficient fine-tuning (PEFT) has become a standard approach for adapting large language models to downstream tasks.
However, most existing PEFT methods rely on uniform and static adaptations, without accounting for the structured heterogeneity of attention across dimensions, heads, layers, and input tokens.
In practice, attention representations exhibit non-uniform behavior, and positional encoding mechanisms such as rotary positional embeddings (RoPE) induce dimension-dependent positional structure, making uniform adaptation suboptimal.
In this work, we propose DyPAM (Dynamic Positional Attention Modulation), a PEFT method that adapts how positional information contributes to attention by operating directly on the query and key representations.
DyPAM combines input-conditioned, dimension-wise modulation with head-wise and layer-wise structural modulation, performing fine-grained adaptation of positional attention aligned with the RoPE-induced structure without modifying the pretrained backbone.
Extensive experiments on mathematical and commonsense reasoning benchmarks across multiple backbone models demonstrate that DyPAM consistently outperforms existing strong PEFT baselines.
The code is available for reproducibility\footnote{\codelink}. 
\end{abstract}

\begin{CCSXML}
<ccs2012>
   <concept>
       <concept_id>10002951.10003227.10003351</concept_id>
       <concept_desc>Information systems~Data mining</concept_desc>
       <concept_significance>500</concept_significance>
       </concept>
   <concept>
       <concept_id>10010147.10010178</concept_id>
       <concept_desc>Computing methodologies~Artificial intelligence</concept_desc>
       <concept_significance>500</concept_significance>
       </concept>
 </ccs2012>
\end{CCSXML}

\ccsdesc[500]{Information systems~Data mining}
\ccsdesc[500]{Computing methodologies~Artificial intelligence}
\keywords{Large Language Models, Parameter-Efficient Fine-Tuning, Positional Attention, Attention Modulation}



\maketitle

\newcommand\kddavailabilityurl{https://doi.org/10.5281/zenodo.20507186}
\ifdefempty{\kddavailabilityurl}{}{
\begingroup\small\noindent\raggedright\textbf{Resource Availability:}\\
% please change the following context to include multiple artifacts if necessary, including data, models, code, etc.
The source code of this paper has been made publicly available at \url{\kddavailabilityurl}.
\endgroup
}

\section{Introduction}

Large language models (LLMs) have been widely adopted across a broad range of real-world applications, including question answering, code generation, and mathematical reasoning~\cite{wang2023rethinking,wu2026icad,cheng2025poi,ma2025spatio}.
Despite their strong general-purpose capabilities, different application scenarios and downstream tasks often impose substantially different requirements on model behavior.
Effectively adapting an LLM to diverse task-specific requirements is essential in practical deployment~\cite{zhao2023survey,li2022eliteplm,li2023web,yu2025bigcity}.

Full-parameter fine-tuning for a specific task is effective but computationally and storage intensive, which limits its practicality in many settings.
Parameter-efficient fine-tuning (PEFT) addresses this issue by updating only a small subset of parameters while keeping the backbone model frozen, achieving a favorable balance between efficiency and performance~\cite{peft,han2024parameter,pan2025contextual,li2025approximation}.
Representative PEFT methods include LoRA\cite{hu2021lora}, which reparameterizes weight updates into a low-rank subspace, IA$^3$~\cite{liu2022few}, which modulates intermediate activations via lightweight scaling vectors, BOFT~\cite{liu2023parameter}, which applies orthogonality-constrained multiplicative transformations with butterfly factorization, and Bone~\cite{kang2024balancing}, which introduces block-wise affine updates with shared parameters.

Despite their strong empirical performance, most existing PEFT methods apply additional parameters in a largely uniform manner across layers, attention heads, and feature dimensions (\ie individual dimensions of the hidden representations), and therefore lack mechanisms to account for the structured and heterogeneous roles of different model components in a fine-grained manner.

\begin{figure}[t]
    \centering
    \includegraphics[width=1\linewidth]{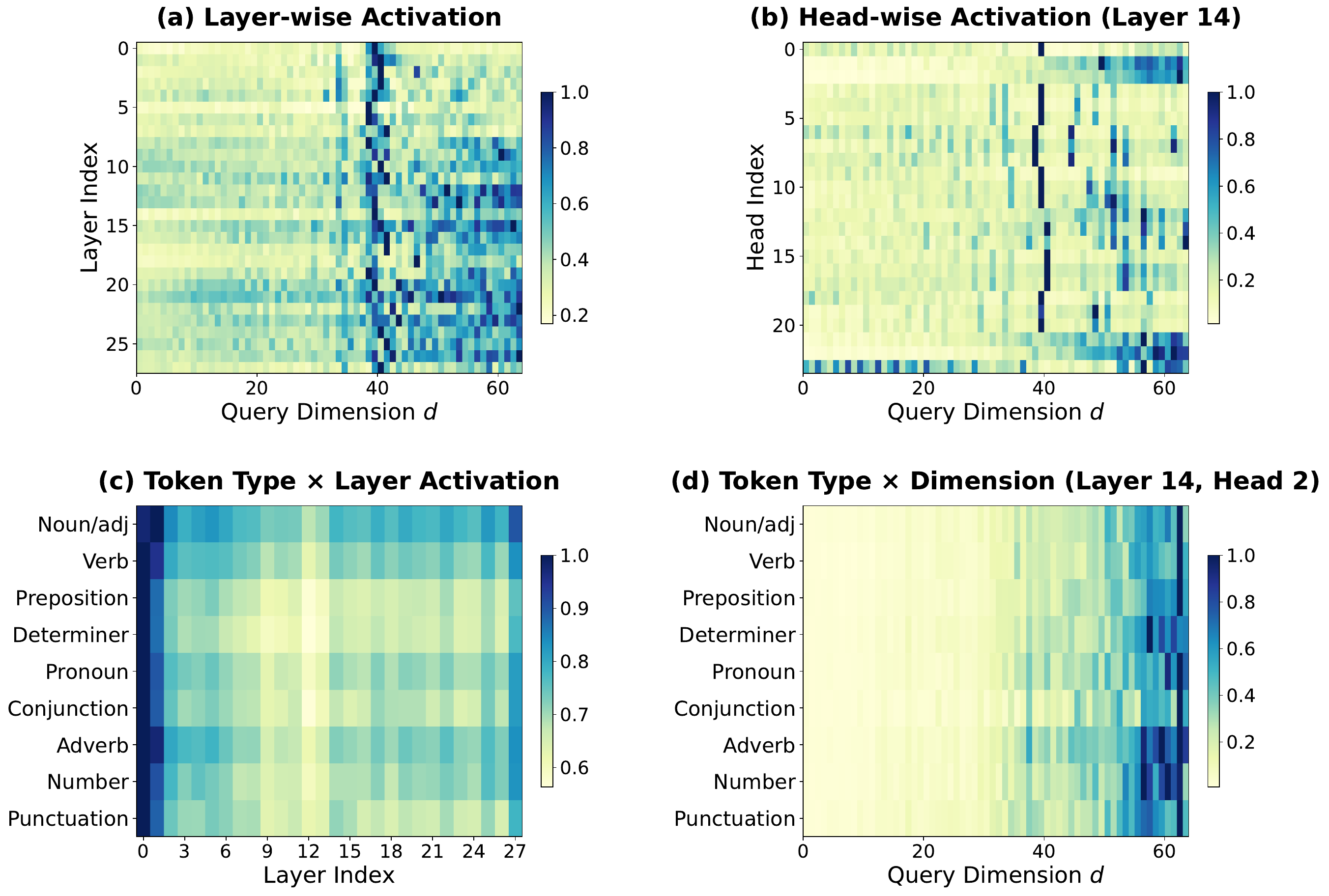}
\caption{Activation heterogeneity in a pretrained Llama-3.2-3B model. The x-axis in (a), (b), and (d) indexes query dimensions of attention mechanism. Activation patterns vary across layers (a), heads (b), and input token types (c, d), indicating that attention operates heterogeneously across dimensions, heads, layers, and tokens.}
    \label{fig:rope_channel}
\end{figure}

Figure~\ref{fig:rope_channel} visualizes attention activation patterns of Llama-3.2-3B, focusing on the query (Q) representations used in self-attention.
Similar heterogeneous patterns are also observed for key (K) representations, while such patterns are not significant for value (V) representations.
Activations vary substantially across query dimensions, attention heads, and layers.
As shown in Figure~\ref{fig:rope_channel}(a), attention exhibits distinct activation patterns across dimensions at different layers, indicating that different dimensions contribute differently to attention.
Figure~\ref{fig:rope_channel}(b) further shows clear head-wise variation within the same layer, indicating that different attention heads exhibit distinct activation patterns.
Figures~\ref{fig:rope_channel}(c) and (d) show that activation patterns also depend on the input tokens.
Tokens with different semantic roles induce systematically different activation distributions, both when aggregated across dimensions and when examined at the level of individual dimensions for a fixed head and layer.
Together, these visualizations demonstrate that attention in pretrained LLMs operates in a structured and heterogeneous manner across dimensions, heads, layers, and input tokens.

Such heterogeneous activation patterns are consistent with the fact that different components of LLMs play distinct functional roles~\cite{zhang2023adalora}.
Recent studies have shown that Transformer modules are not functionally uniform, with different layers and components specializing in different aspects~\cite{geva2021transformer}.
For example, feed-forward networks have been shown to primarily store factual information~\cite{meng2022mass}, while attention mechanisms are more closely associated with contextual interaction~\cite{clark2019does}.
Within the attention module, architectural design choices further introduce structured differences in how information is processed.
In particular, positional encoding mechanisms in Transformer-based LLMs introduce structured differences in internal representations, resulting in non-uniform usage of attention dimensions across layers and heads~\cite{barbero2024round,jin2025massive}.
Beyond structural differences across model components, attention behavior also varies with the input, where different tokens can induce different activation patterns depending on their semantic roles and contextual requirements~\cite{wang2023wavelet,ji2026break}.
Taken together, these findings suggest that LLMs exhibit intrinsic functional heterogeneity across components and inputs, and effective adaptation methods should account for such fine-grained structural differences.

These observations suggest three requirements for structure-aware adaptation.
First, adaptation should distinguish attention dimensions, since different dimensions exhibit different positional and functional behaviors.
Second, it should account for head-wise and layer-wise differences, because the same dimension can play different roles across attention heads and network depths.
Third, adaptation should remain input-conditioned, as different tokens and contexts can induce different attention activation patterns.
This provides a direct way to adapt dependency modeling while keeping the original attention architecture unchanged.

Motivated by these observations, we propose \textbf{DyPAM} (\textbf{Dy}namic \textbf{P}ositional \textbf{A}ttention \textbf{M}odulation), a PEFT method that adapts how positional information contributes to attention.
DyPAM addresses attention heterogeneity from two complementary perspectives.
It combines input-conditioned, dimension-wise modulation with head-wise and layer-wise structural modulation to adapt attention behavior.
In contrast to prior PEFT methods that apply static adaptation shared across inputs, DyPAM models heterogeneity jointly at the token, dimension, head, and layer levels, and aligns the modulation with the RoPE-induced dimension pairing.
Extensive experiments across multiple models and tasks demonstrate the effectiveness of this design.

We summarize the main contributions of this work as follows:
\begin{itemize}[leftmargin=*, topsep=0pt]
    \item We propose \textbf{DyPAM}, the first parameter-efficient fine-tuning method that explicitly adapts large language models by modulating positional attention in a fine-grained and structured manner.
    \item DyPAM introduces \emph{input-conditioned, dimension-wise modulation}, enabling different attention dimensions to be dynamically adjusted according to the input context.
    \item DyPAM further incorporates \emph{head-wise and layer-wise structural modulation}, allowing different attention heads and network layers to maintain distinct positional preferences.
    \item Extensive experiments across multiple backbone models and downstream tasks demonstrate the effectiveness and robustness of DyPAM compared to existing PEFT methods.
\end{itemize}

\section{Preliminaries}
We briefly review the attention mechanism, Rotary Position Embedding (RoPE), and parameter-efficient fine-tuning (PEFT), which provide the necessary background for describing our method.
\subsection{Attention in Large Language Models}
\label{sec:prelim_attention}
Large language models are typically built upon the Transformer architecture~\cite{vaswani2017attention}, where self-attention is the core mechanism for modeling interactions among tokens.
Given an input sequence, each layer $\ell$ of the model produces a sequence of hidden states
$\mathbf{H}^{(\ell)} = [\mathbf{h}^{(\ell)}_1, \mathbf{h}^{(\ell)}_2, \dots, \mathbf{h}^{(\ell)}_T]$,
where $\mathbf{h}^{(\ell)}_t \in \mathbb{R}^{d}$ denotes the hidden representation of the $t$-th token.
In each attention layer, the hidden states are linearly projected into query, key, and value representations,
\begin{equation}
\mathbf{Q}^{(\ell)} = \mathbf{H}^{(\ell)} \mathbf{W}_Q^{(\ell)}, \quad
\mathbf{K}^{(\ell)} = \mathbf{H}^{(\ell)} \mathbf{W}_K^{(\ell)}, \quad
\mathbf{V}^{(\ell)} = \mathbf{H}^{(\ell)} \mathbf{W}_V^{(\ell)},
\label{eq:hk_to_qk}
\end{equation}
where $\mathbf{W}_Q^{(\ell)}, \mathbf{W}_K^{(\ell)}, \mathbf{W}_V^{(\ell)} \in \mathbb{R}^{d \times d}$ are learned projection matrices.

The projected representations are then reshaped into $H$ attention heads.
For each head $h \in \{1,\dots,H\}$, we denote the per-head matrices as
$\mathbf{Q}^{(\ell,h)}, \mathbf{K}^{(\ell,h)}, \mathbf{V}^{(\ell,h)} \in \mathbb{R}^{T \times d_{\text{head}}}$,
where $d_{\text{head}} = d / H$.
For a given token position $t$, we denote by
$\mathbf{q}^{(\ell,h)}_t \in \mathbb{R}^{d_{\text{head}}}$ and
$\mathbf{k}^{(\ell,h)}_t \in \mathbb{R}^{d_{\text{head}}}$
the query and key vectors corresponding to the $t$-th token.
We further denote by $q^{(\ell,h)}_{t,j}$ the $j$-th feature dimension of $\mathbf{q}^{(\ell,h)}_t$.
Self-attention is computed independently for each head.
For head $h$ at layer $\ell$, the attention output is given by
\begin{equation}
\mathrm{Attn}\!\left(\mathbf{Q}^{(\ell,h)}, \mathbf{K}^{(\ell,h)}, \mathbf{V}^{(\ell,h)}\right)
=
\mathrm{softmax}\!\left(
\frac{\mathbf{Q}^{(\ell,h)} {\mathbf{K}^{(\ell,h)}}^\top}{\sqrt{d_{\text{head}}}}
\right)\mathbf{V}^{(\ell,h)}.
\end{equation}

Notably, the attention operation itself is permutation-invariant and does not encode positional order~\cite{press2021train}.
As a result, positional information must be explicitly incorporated into the attention computation.
Modern LLMs achieve this by applying position-dependent transformations to the query and key representations.

\begin{figure}[t]
    \centering
    \includegraphics[width=1\linewidth]{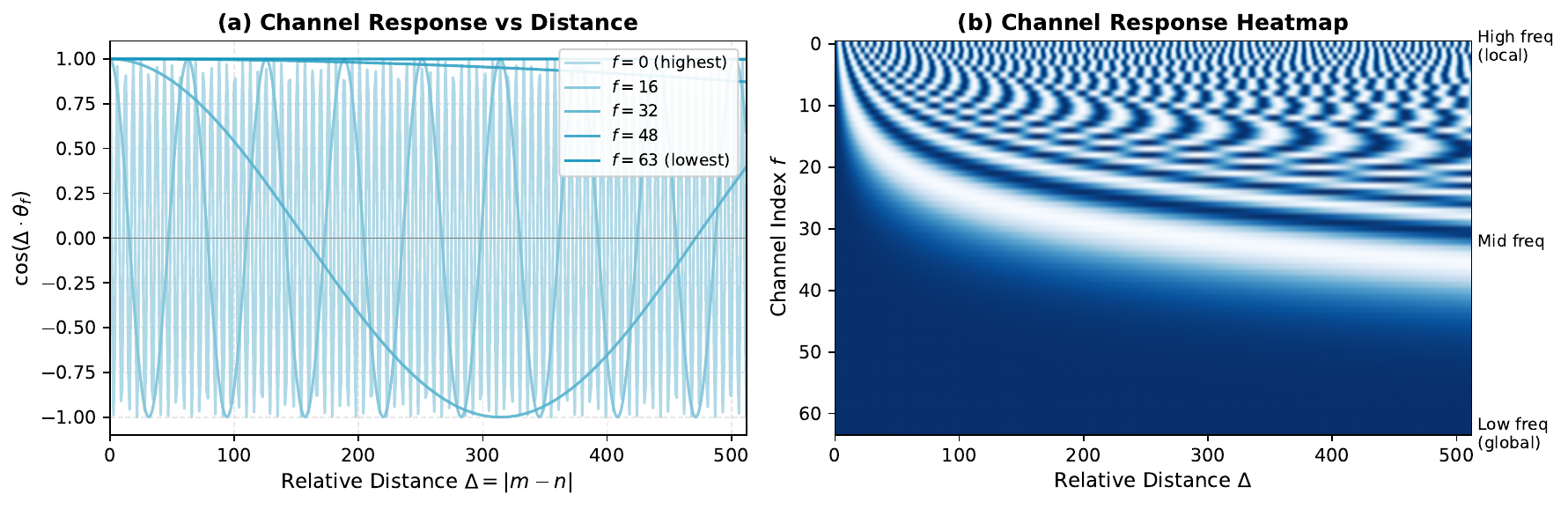}
\caption{
Position-dependent responses across attention dimensions induced by RoPE.
(a) Different dimensions respond differently to relative positional distances.
(b) Heatmap of all dimensions, showing non-uniform positional sensitivity.
}
    \label{fig:rope_math}
\end{figure}

\begin{figure*}[ht]
    \centering
    \includegraphics[width=0.9\linewidth]{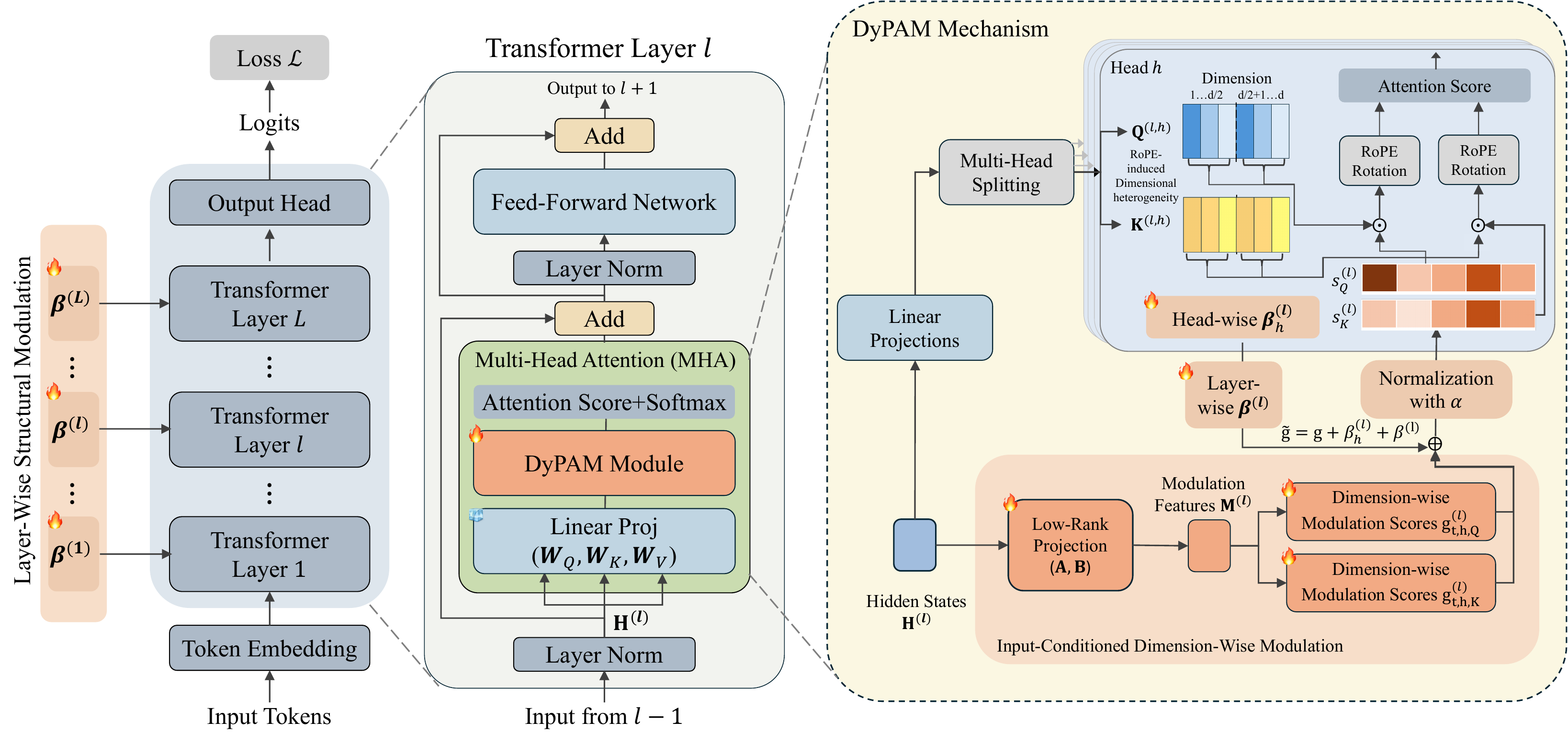}
    \caption{The architecture of DyPAM framework. DyPAM applies input-conditioned, dimension-wise modulation together with head-wise and layer-wise structural biases to the query and key representations before RoPE, enabling fine-grained adaptation of positional attention within the PEFT paradigm.}
    \label{fig:framework}
\end{figure*}
\subsection{Rotary Position Embedding}

Rotary Position Embedding (RoPE)~\cite{su2024roformer} incorporates positional information into attention by applying position-dependent transformations to the query and key representations.
For each attention head, RoPE views the per-head vector as two halves of equal size: a ``real'' part and an ``imaginary'' part, each of dimension $d_{\text{head}}/2$.
RoPE then applies a 2D rotation to each index $i$ by jointly rotating the paired components
$\big(z^{\text{real}}_i, z^{\text{imag}}_i\big)$.
As a consequence, the two halves share the same rotation at each index, and thus exhibit closely related positional behavior across corresponding dimensions.

The rotations vary across dimensions, resulting in different positional behaviors.
Figure~\ref{fig:rope_math} visualizes this effect, showing that different attention dimensions respond differently to relative positional distances in a non-uniform and multi-scale manner.
As a result, different attention dimensions play different roles in encoding positional information, which motivates treating them differently when adapting positional attention.

\subsection{Parameter-Efficient Fine-Tuning}

Parameter-efficient fine-tuning (PEFT) adapts LLMs by introducing lightweight transformations, while keeping the pretrained backbone fixed.
We describe PEFT by applying a constrained transformation to an intermediate representation $\mathbf{z}$ in the model.
Formally, the adapted representation can be written as
\begin{equation}
\mathbf{z}' = \mathcal{T}\big(\mathbf{z}; \boldsymbol{\theta}\big),
\end{equation}
where $\mathcal{T}(\cdot;\boldsymbol{\theta})$ denotes a parameter-efficient transformation with a small number of trainable parameters $\boldsymbol{\theta}$.
Many existing PEFT methods adopt an additive update,
\begin{equation}
\mathbf{z}' = \mathbf{z} + \Delta \mathbf{z},
\qquad
\Delta \mathbf{z} = \mathcal{A}\big(\mathbf{z}; \boldsymbol{\theta}\big),
\end{equation}
where the adaptation module $\mathcal{A}(\cdot)$ is typically applied in a static and largely uniform manner across model components.

While effective, this formulation typically applies the same adaptation mechanism uniformly across layers, attention heads, feature dimensions, and input tokens.
However, attention behavior in modern LLMs is highly heterogeneous, particularly in how positional information is encoded.
This motivates PEFT approaches that perform fine-grained and structure-aware adaptation.

In this work, we follow the PEFT paradigm and apply multiplicative modulation to attention representations:
\begin{equation}
\mathbf{z}' = \mathbf{s}(\mathbf{x}) \odot \mathbf{z},
\qquad
\mathbf{s}(\mathbf{x}) = \mathcal{M}\big(\mathbf{x}; \boldsymbol{\theta}\big),
\label{eq:mypeft}
\end{equation}
where $\mathbf{s}(\mathbf{x})$ is an input-conditioned modulation signal and $\odot$ denotes element-wise multiplication.
In DyPAM, this modulation is applied to the query and key representations in attention and indexed by the internal structure of attention, including layers, heads, tokens, and feature dimensions.
Both formulations follow the same parameter-efficient adaptation paradigm.
The key difference is that DyPAM performs input-conditioned, structure-aware modulation, enabling fine-grained adaptation of positional attention.

\section{Method}
\label{sec:method}

In this section, we introduce DyPAM, a PEFT method for LLMs that explicitly modulates positional information in attention. We first present an overview of the DyPAM framework and its core design principles, and then describe each component in detail.
\subsection{Framework Overview}

In RoPE-based LLMs, attention exhibits heterogeneous behavior across feature dimensions, layers, heads, and input tokens.
However, most existing PEFT methods rely on uniform and static adaptation mechanisms, without accounting for such heterogeneity.

To address this limitation, we propose \textbf{Dynamic Positional Attention Modulation (DyPAM)}, a parameter-efficient fine-tuning method that adapts how positional information contributes to attention.
DyPAM operates directly on the query and key representations and jointly models input-conditioned, dimension-wise modulation together with head-wise and layer-wise structural modulation, enabling fine-grained and structured adaptation of positional attention.
As illustrated in Figure~\ref{fig:framework}, DyPAM introduces modulation into attention representations in a manner aligned with the internal structure of attention and conditioned on the input.
The two modulation sources play different roles.
The input-conditioned component captures token-dependent variation by generating modulation features from hidden states.
The structural bias terms capture persistent preferences associated with heads and layers.
Their combination allows DyPAM to adapt to each input while maintaining stable structural differences across the attention hierarchy.

In the following, we present DyPAM by detailing how the modulation is constructed and how it is applied to attention representations, and summarize the overall procedure in Algorithm~\ref{alg:dypam}.
\subsection{Query--Key Representations and Modulation Features}
\label{sec:modulation_feature}

DyPAM operates on the query and key representations used in self-attention.
At each Transformer layer $\ell$, these representations are derived from the token-level hidden states
$\mathbf{H}^{(\ell)} \in \mathbb{R}^{B \times T \times d}$, where $B$ is the batch size, $T$ is the sequence length, and $d$ is the model dimension.
Following the standard attention formulation described in Section~\ref{sec:prelim_attention},
the hidden states are linearly projected to obtain the query and key matrices
$\mathbf{Q}^{(\ell)}$ and $\mathbf{K}^{(\ell)}$,
which are subsequently reshaped into per-head representations
$\mathbf{Q}^{(\ell,h)}$ and $\mathbf{K}^{(\ell,h)}$ as defined in Eq.~\eqref{eq:hk_to_qk}.

To enable input-conditioned adaptation of attention behavior, DyPAM derives modulation features directly from the same hidden states $\mathbf{H}^{(\ell)}$.
Since the hidden states encode token-specific contextual information, the resulting modulation features are token-dependent and differ across inputs, providing the basis for input-conditioned modulation.
Concretely, we apply a lightweight low-rank projection to the hidden states, yielding modulation features:
\begin{equation}
\mathbf{M}^{(\ell)} = \mathbf{H}^{(\ell)} \mathbf{A}^{(\ell)} \mathbf{B}^{(\ell)},
\qquad
\mathbf{M}^{(\ell)} \in \mathbb{R}^{B \times T \times (H \cdot d_e)},
\label{eq:modulation_feature}
\end{equation}
where $\mathbf{A}^{(\ell)} \in \mathbb{R}^{d \times r}$ and
$\mathbf{B}^{(\ell)} \in \mathbb{R}^{r \times (H \cdot d_e)}$ are learnable matrices with rank $r \ll d$,
$H$ is the number of attention heads, and $d_e$ denotes the feature dimension per head.

The projected features are reshaped into $H$ head-specific components,
yielding modulation features $\mathbf{m}^{(\ell)}_{t,h} \in \mathbb{R}^{d_e}$ for each token position $t$ and attention head $h$.
These features encode contextual information associated with each token and attention head, capturing how the current input is represented at different heads within the layer.
They serve as an intermediate representation that bridges the token-level hidden states and the dimension-wise modulation subsequently applied to the query and key representations.

\subsection{Input-Conditioned Dimension-Wise Modulation}
\label{sec:dim_modulation}

Given the modulation features constructed from the hidden states, DyPAM maps them to dimension-wise modulation values that are aligned with the query and key representations in attention.
This mapping determines how the contribution of each attention dimension is modulated in an input-conditioned manner, allowing the model to dynamically adjust the influence of each dimension.
By conditioning on both the token and its context, DyPAM enables fine-grained control over how positional information is utilized across different attention dimensions.

For each layer $\ell$, DyPAM introduces learnable dimension embedding matrices that project modulation features to the attention dimension space.
Concretely, for the query and key representations, we use separate embedding matrices
\begin{equation}
\mathbf{E}^{(\ell)}_Q \in \mathbb{R}^{\frac{d_{\text{head}}}{2} \times d_e},
\qquad
\mathbf{E}^{(\ell)}_K \in \mathbb{R}^{\frac{d_{\text{head}}}{2} \times d_e},
\label{eq:dim_embedding}
\end{equation}
where each row corresponds to one pair of attention dimensions.
This design reflects the structure induced by RoPE, where each pair of dimensions shares the same positional rotation and therefore exhibits similar positional behavior.
By assigning a single modulation value to each dimension pair, DyPAM reduces parameter overhead while respecting the RoPE-induced structure.
This pair-wise sharing also avoids assigning unrelated modulation values to two coordinates that jointly represent the same rotated subspace.
In this way, DyPAM changes the strength of each RoPE-aligned channel without breaking the internal pairing used by the positional transformation.
This also supports grouped-query attention (GQA)~\cite{ainslie2023gqa}, where multiple attention heads share key and value projections. 
In such cases, the key-side modulation is shared across heads that use the same key representation, while the query-side modulation remains head-specific.

Given the modulation feature $\mathbf{m}^{(\ell)}_{t,h} \in \mathbb{R}^{d_e}$ for token position $t$ and attention head $h$, the dimension-wise modulation scores for queries and keys are computed as
\begin{equation}
\mathbf{g}^{(\ell)}_{t,h,Q}
=
\mathbf{E}^{(\ell)}_Q \mathbf{m}^{(\ell)}_{t,h},
\qquad
\mathbf{g}^{(\ell)}_{t,h,K}
=
\mathbf{E}^{(\ell)}_K \mathbf{m}^{(\ell)}_{t,h},
\label{eq:dim_score}
\end{equation}
where
$\mathbf{g}^{(\ell)}_{t,h,Q}, \mathbf{g}^{(\ell)}_{t,h,K} \in \mathbb{R}^{\frac{d_{\text{head}}}{2}}$
denote the modulation scores for query and key dimension pairs, respectively.
Each entry corresponds to one RoPE dimension pair, and we use $i$ to index dimension pair $i$.

At this stage, the modulation scores provide input-conditioned, dimension-wise adjustments for the query and key representations.

\paratitle{Discussion.}
Input-conditioned dimension-wise modulation enables DyPAM to adapt the contribution of individual attention dimensions based on the input context. By aligning modulation with RoPE-induced dimension pairing, DyPAM selectively adjusts how positional information influences attention, while maintaining parameter efficiency. This mechanism provides fine-grained control over positional attention that is sensitive to both token-level context and the structured organization of attention dimensions.

\subsection{Head-Wise and Layer-Wise Structural Modulation}
\label{sec:structural_modulation}

While input-conditioned dimension-wise modulation captures token-dependent variation, attention behavior also exhibits differences across attention heads and network layers.
To model such structure-level heterogeneity, DyPAM introduces head-wise and layer-wise structural modulation that is independent of the input.

For each layer $\ell$, DyPAM maintains layer-wise bias vectors
\begin{equation}
\boldsymbol{\beta}^{(\ell)}_{Q}, \boldsymbol{\beta}^{(\ell)}_{K} \in \mathbb{R}^{\frac{d_{\text{head}}}{2}},
\end{equation}
which capture layer-specific preferences over attention dimension pairs for the query and key representations.
In addition, for each attention head $h$ at layer $\ell$, DyPAM introduces head-wise bias vectors
\begin{equation}
\boldsymbol{\beta}^{(\ell)}_{h,Q}, \boldsymbol{\beta}^{(\ell)}_{h,K} \in \mathbb{R}^{\frac{d_{\text{head}}}{2}},
\end{equation}
allowing different heads within the same layer to maintain distinct structural biases.
Here the subscripts $Q$ and $K$ indicate that the query and key representations use separate bias vectors of the same shape.
These bias terms are added to the dimension-wise modulation scores.
For queries and keys, the structurally augmented modulation scores are given by
\begin{equation}
\tilde{\mathbf{g}}^{(\ell)}_{t,h,Q}
=
\mathbf{g}^{(\ell)}_{t,h,Q}
+
\boldsymbol{\beta}^{(\ell)}_{h,Q}
+
\boldsymbol{\beta}^{(\ell)}_{Q},
\label{structural_modulation_Q}
\end{equation}
\begin{equation}
\tilde{\mathbf{g}}^{(\ell)}_{t,h,K}
=
\mathbf{g}^{(\ell)}_{t,h,K}
+
\boldsymbol{\beta}^{(\ell)}_{h,K}
+
\boldsymbol{\beta}^{(\ell)}_{K},
\label{structural_modulation_K}
\end{equation}
where $\mathbf{g}^{(\ell)}_{t,h,Q}$ and $\mathbf{g}^{(\ell)}_{t,h,K}$ are the input-conditioned dimension-wise scores from Section~\ref{sec:dim_modulation}.
The bias terms are shared across token positions and encode structural preferences that persist across inputs.

At this stage, the modulation scores integrate input-conditioned, dimension-wise adjustments with head-wise and layer-wise structural biases, capturing both token-dependent variation and persistent structural preferences in attention.
These scores are then transformed into bounded modulation factors and applied to the query and key representations.

\begin{algorithm}[t]
\caption{DyPAM: Dynamic Positional Attention Modulation}
\label{alg:dypam}
\begin{algorithmic}[1]
\Require
Input sequence $\mathbf{x} = (x_1, \dots, x_T)$, pretrained RoPE-based LLM, DyPAM parameters
\Ensure
Model output distribution and training loss $\mathcal{L}$

\State Obtain token embeddings from $\mathbf{x}$

\For{each Transformer layer $\ell = 1, \dots, L$}
    \State Compute hidden states $\mathbf{H}^{(\ell)}$

    \State Project hidden states to query and key representations
    \State \hspace{1em} $\mathbf{Q}^{(\ell)}, \mathbf{K}^{(\ell)}$ according to Eq.~\eqref{eq:hk_to_qk}
    \State Reshape $\mathbf{Q}^{(\ell)}, \mathbf{K}^{(\ell)}$ into per-head representations
    \State \hspace{1em} $\mathbf{Q}^{(\ell,h)}, \mathbf{K}^{(\ell,h)}$

    \State Construct modulation features from hidden states
    \State \hspace{1em} $\mathbf{m}^{(\ell)}_{t,h}$ according to Eq.~\eqref{eq:modulation_feature}

    \State Compute input-conditioned, dimension-wise modulation scores
    \State \hspace{1em} $\mathbf{g}^{(\ell)}_{t,h,Q}, \mathbf{g}^{(\ell)}_{t,h,K}$ according to Eq.~\eqref{eq:dim_score}

    \State Add head-wise and layer-wise structural biases
    \State \hspace{1em} $\tilde{\mathbf{g}}^{(\ell)}_{t,h,Q}, \tilde{\mathbf{g}}^{(\ell)}_{t,h,K}$ according to Eq.~\eqref{structural_modulation_Q} and \eqref{structural_modulation_K}

    \State Normalize modulation scores to obtain modulation factors
    \State \hspace{1em} $s^{(\ell)}_{t,h,i}$ according to Eq.~\eqref{eq:modulation_factor}

    \State Apply modulation to query and key representations
    \State \hspace{1em} $\hat{\mathbf{Q}}^{(\ell,h)}, \hat{\mathbf{K}}^{(\ell,h)}$ according to Eq.~\eqref{eq:apply_qk}

    \State Apply RoPE to modulated query and key representations

    \State Compute attention outputs using modulated queries and keys
\EndFor

\State Compute model outputs and training loss $\mathcal{L}$ using Eq.~\eqref{eq:training_loss}

\end{algorithmic}
\end{algorithm}

\subsection{Applying Modulation in Attention}
\label{sec:apply_modulation}

The combined modulation scores obtained from the previous steps encode both input-conditioned and structural adjustments over attention dimensions.
We next apply a normalization step that maps these scores to bounded modulation factors, ensuring stable and controlled adaptation.

Here, $\tilde{g}^{(\ell)}_{t,h,i}$ denotes the modulation score for dimension pair $i$ in the combined score vector for the query or key representation at layer $\ell$, head $h$, and token position $t$.
For brevity, we omit the explicit $Q/K$ subscript when the formulation applies identically to both.
For each layer $\ell$, token position $t$, attention head $h$, and dimension pair $i$, the normalized modulation factor is computed as
\begin{equation}
s^{(\ell)}_{t,h,i}
=
1 + \alpha \cdot \big(\sigma(\tilde{g}^{(\ell)}_{t,h,i}) - 0.5\big),
\label{eq:modulation_factor}
\end{equation}
where $\sigma(\cdot)$ is the sigmoid function and $\alpha$ controls the modulation strength.
This normalization maps the modulation factors to a bounded interval
$
\big[1 - \tfrac{\alpha}{2},\, 1 + \tfrac{\alpha}{2}\big]
$,
centering them around the original scale and preventing deviation from the pretrained representations.

The modulation factors are applied to the query and key representations before positional encoding.
Let $\mathbf{q}^{(\ell,h)}_{t,i} \in \mathbb{R}^{2}$ and
$\mathbf{k}^{(\ell,h)}_{t,i} \in \mathbb{R}^{2}$
denote the paired dimensions of the query and key vectors corresponding to dimension pair $i$.
Both dimensions within each pair are modulated using the same factor:
\begin{equation}
\hat{\mathbf{q}}^{(\ell,h)}_{t,i}
=
s^{(\ell)}_{t,h,i} \cdot \mathbf{q}^{(\ell,h)}_{t,i},
\qquad
\hat{\mathbf{k}}^{(\ell,h)}_{t,i}
=
s^{(\ell)}_{t,h,i} \cdot \mathbf{k}^{(\ell,h)}_{t,i}.
\label{eq:apply_qk}
\end{equation}

This operation corresponds to the multiplicative PEFT formulation introduced in Eq.~\eqref{eq:mypeft},
where the pretrained representation $\mathbf{z}$ corresponds to the query or key vectors,
and the modulation signal $\mathbf{s}(\mathbf{x})$ is indexed by layer, head, token position, and dimension pair.
Following the convention above, $s^{(\ell)}_{t,h,i}$ is computed independently for the query and key sides from their respective scores $\tilde{\mathbf{g}}^{(\ell)}_{t,h,Q}$ and $\tilde{\mathbf{g}}^{(\ell)}_{t,h,K}$, so the query and key representations are modulated by their own factors.
As the two coordinates within each pair share the same factor, the modulation signal $\mathbf{s}(\mathbf{x})$ takes a common value over each pair and thus spans the full $d_{\text{head}}$ dimensions of the query and key vectors.
The modulated query and key representations are then passed through the RoPE mechanism and used in the standard attention computation.
By applying modulation prior to RoPE, DyPAM aligns adaptation with the RoPE-induced positional structure.

More broadly, DyPAM addresses attention heterogeneity by jointly modeling input-conditioned, dimension-wise modulation and head-wise, layer-wise structural modulation.
This design enables attention dimensions to adapt to input context, while allowing different heads and layers to maintain distinct positional preferences.
Rather than introducing uniform parameter updates, DyPAM performs targeted modulation aligned with how positional information is encoded and utilized in attention.
As a result, DyPAM provides a structured and fine-grained mechanism for adapting positional attention within the PEFT paradigm.

\subsection{Training Details}
\label{sec:training}

DyPAM is trained end-to-end using the standard cross-entropy loss for language modeling.
Given an input sequence $\mathbf{x} = (x_1, \dots, x_T)$ and the corresponding target sequence $\mathbf{y} = (y_1, \dots, y_T)$, the training loss is defined as
\begin{equation}
\mathcal{L}
=
-\sum_{t=1}^{T} \log p(y_t \mid x_{\leq t}),
\label{eq:training_loss}
\end{equation}
where $p(y_t \mid x_{\leq t})$ denotes the model output distribution at position $t$.
The model is parameterized by the pretrained backbone together with the DyPAM parameters.
Algorithm~\ref{alg:dypam} summarizes the overall forward computation and training procedure.

\begin{table*}[t]
    \centering
    \small
\caption{Comparison of \framework with PEFT baselines on mathematical reasoning benchmarks across three backbone models.
% Results are reported in accuracy (\%).
Micro-avg and macro-avg denote micro- and macro-averaged performance. Best results are highlighted in bold, and second-best results are underlined.
$^{*}$ indicates statistically significant improvements over the best baseline (two-sided t-test, $p<0.05$).}
    \resizebox{1\linewidth}{!}{
        \renewcommand{\arraystretch}{0.95}
        \begin{tabular}{l|l ccccccc ccc}
            \toprule
            \textbf{Backbone LLM} & \textbf{Method}
            & \textbf{Param(\%)}
            & \textbf{MultiArith} & \textbf{GSM8K} & \textbf{AddSub}
            & \textbf{AQuA} & \textbf{SingleEq}
            & \textbf{SVAMP} & \textbf{MAWPS}
            & \textbf{micro-avg(\%)$\uparrow$}
            & \textbf{macro-avg(\%)$\uparrow$} \\
            \midrule
            \multirow{10}{*}{\textbf{LLaMA 3.2 3B}}
                & LoRA      & 1.12 & 71.50 & 33.21 & 78.48 & 22.44 & 81.50 & 54.10 & 76.47 & 54.96 & 59.67 \\
                & AdaLoRA   & 2.22 & 75.67 & 36.32 & 80.51 & 22.83 & 87.80 & 55.60 & 78.57 & 57.90 & 62.47 \\
                & OFT       & 0.73 & 87.17 & \textbf{40.18} & \underline{85.82} & \underline{24.02} & 86.42 & 61.50 & \textbf{84.03} & 62.75 & \underline{67.02} \\
                & Bone      & 1.14 & \underline{87.50} & 39.73 & 85.57 & 23.62 & 86.61 & \textbf{63.70} & 81.93 & \underline{63.03} & 66.95 \\
                & IA$^3$    & 0.02 & 58.33 & 27.37 & 68.61 & 20.47 & 72.83 & 47.90 & 58.82 & 46.89 & 50.62 \\
                & LN-Tuning & 0.01 & 58.00 & 26.38 & 66.58 & 21.26 & 74.80 & 44.90 & 60.08 & 46.01 & 50.29 \\
                & FourierFT & 0.73 & 78.67 & 33.21 & 82.03 & 20.47 & 85.43 & 54.30 & 77.31 & 56.72 & 61.63 \\
                & SHiRA     & 1.12 & 82.50 & 38.82 & 84.81 & \underline{24.02} & \underline{87.99} & 56.90 & 81.93 & 60.59 & 65.28 \\
                & RoSA      & 0.54 & 84.33 & 37.91 & 82.78 & 22.83 & 87.01 & 52.50 & 78.99 & 59.02 & 63.77 \\
                & \framework (ours)
                             & 0.92 & \textbf{88.50} & \underline{39.88} & \textbf{86.33}
                             & \textbf{25.20} & \textbf{88.78} & \underline{63.00} & \textbf{84.03}
                             & \textbf{63.58}$^{*}$ & \textbf{67.96*} \\
            \midrule
            \multirow{10}{*}{\textbf{Qwen3 8B}}
                & LoRA      & 0.79 & 97.67 & 74.91 & 89.87 & 35.83 & 90.55 & 84.70 & 89.08 & 82.04 & 80.37 \\
                & AdaLoRA   & 1.57 & 95.17 & 73.01 & 90.63 & 37.01 & 92.32 & 84.80 & 91.60 & 81.62 & 80.65 \\
                & OFT       & 0.51 & 95.67 & 73.46 & 90.38 & 33.07 & 94.09 & 84.90 & 91.60 & 81.80 & 80.45 \\
                & Bone      & 0.81 & \underline{98.00} & 72.25 & \underline{91.65} & 33.46 & 93.90 & 83.80 & 90.34 & 81.55 & 80.49 \\
                & IA$^3$    & 0.02 & 92.50 & 72.18 & 84.81 & 35.04 & 86.61 & 80.90 & 86.55 & 78.49 & 76.94 \\
                & LN-Tuning & 0.00 & 91.67 & 68.69 & 85.32 & \underline{39.76} & 87.40 & 78.00 & 85.71 & 77.01 & 76.65 \\
                & FourierFT & 0.37 & 94.50 & 70.05 & 87.34 & 31.50 & 86.81 & 82.70 & 81.09 & 78.28 & 76.28 \\
                & SHiRA     & 0.79 & 94.83 & \underline{75.36} & 90.13 & 37.01 & 93.90 & \textbf{85.70} & 90.34 & \underline{82.57} & 81.04 \\
                & RoSA      & 0.36 & 97.83 & 74.07 & 90.38 & 35.43 & \underline{94.49} & 84.80 & \underline{92.02} & 82.48 & \underline{81.29} \\
                & \framework (ours)
                             & 0.61 & \textbf{99.17} & \textbf{76.72} & \textbf{91.90}
                             & \textbf{40.94} & \textbf{95.28} & \underline{85.50} & \textbf{92.86}
                             & \textbf{84.24}$^{*}$ & \textbf{83.20*} \\
            \midrule
            \multirow{10}{*}{\textbf{Gemma 3 4B}}
                & LoRA      & 1.33 & 86.00 & 51.25 & 72.41 & 25.98 & 75.59 & 62.20 & 75.21 & 63.26 & 64.09 \\
                & AdaLoRA   & 2.62 & 82.67 & 51.86 & 66.33 & 31.50 & 73.82 & 62.30 & 73.95 & 62.49 & 63.20 \\
                & OFT       & 0.75 & 85.83 & \underline{54.28} & 72.91 & \underline{32.28} & 75.59 & \textbf{63.80} & \underline{76.47} & \underline{65.02} & \underline{65.88} \\
                & Bone      & 1.41 &\underline{86.17} & 45.87 & 71.39 & 30.31 & 72.64 & 55.10 & 73.11 & 59.69 & 62.08 \\
                & IA$^3$    & 0.03 & 42.67 & 38.89 & 40.51 & 27.17 & 40.75 & 37.20 & 37.39 & 38.62 & 37.80 \\
                & LN-Tuning & 0.01 & 32.67 & 30.63 & 45.06 & 23.62 & 56.69 & 40.80 & 37.82 & 37.64 & 38.18 \\
                & FourierFT & 1.10 & 60.83 & 31.24 & 65.32 & 28.35 & 66.73 & 46.30 & 65.97 & 47.89 & 52.10 \\
                & SHiRA     & 1.33 & 72.67 & 42.08 & \underline{73.16} & 31.50 & \textbf{76.57} & 61.30 & 75.63 & 58.92 & 61.84 \\
                & RoSA      & 0.40 & 34.50 & 38.51 & 66.84 & 31.10 & 63.19 & 43.70 & 62.18 & 45.53 & 48.58 \\
                & \framework (ours)
                             & 0.62 & \textbf{86.33} & \textbf{55.19} & \textbf{73.42}
                             & \textbf{32.68} & \underline{76.18} & \underline{62.70} & \textbf{76.89}
                             & \textbf{65.28}$^{*}$ & \textbf{66.20*} \\
            \bottomrule
        \end{tabular}
    }
    \label{tab:main_math}
    \vspace{-4px}
\end{table*}

\section{Experiments}
This section evaluates DyPAM across tasks, models, and experimental settings.
Our experimental design is organized around a set of research questions that assess overall performance, scalability across model sizes, component contributions, and learned positional modulation behavior.

Specifically, we investigate the following research questions:
\begin{itemize}[leftmargin=*, topsep=0pt]
    \item \textbf{RQ1:} Does DyPAM outperform existing PEFT methods with comparable numbers of trainable parameters?
    \item \textbf{RQ2:} Is DyPAM effective across different backbone model sizes?
    \item \textbf{RQ3:} Which components of DyPAM contribute most to its performance?
    \item \textbf{RQ4:} How sensitive is DyPAM to its hyperparameters?
    \item \textbf{RQ5:} What positional modulation patterns does DyPAM learn?
\end{itemize}

We first introduce the experimental setup, followed by a systematic evaluation of DyPAM with respect to each research question.

\subsection{Experimental Setup}
\paratitle{Datasets.}
We evaluate DyPAM on mathematical and commonsense reasoning tasks with distinct reasoning patterns and positional sensitivity, and train on two datasets drawn from multiple existing benchmarks~\cite{hu2023llm}, emphasizing multi-step arithmetic and general commonsense reasoning, respectively.
Mathematical reasoning performance is evaluated on GSM8K~\cite{cobbe2021training}, AQuA~\cite{ling2017program}, MAWPS~\cite{koncel2016mawps}, AddSub~\cite{hosseini2014learning}, MultiArith~\cite{roy2016solving}, SingleEq~\cite{koncel2015parsing}, and SVAMP~\cite{patel2021nlp}, while commonsense reasoning is evaluated on BoolQ~\cite{clark2019boolq}, PIQA~\cite{bisk2020piqa}, Social IQA~\cite{sap2019socialiqa}, ARC-Easy and ARC-Challenge~\cite{clark2018think}, OpenBookQA~\cite{mihaylov2018can}, HellaSwag~\cite{zellers2019hellaswag}, and WinoGrande~\cite{sakaguchi2020winogrande}. 
Across benchmarks, inputs are natural-language questions and outputs are either final numeric answers or discrete choices, and we report accuracy as the evaluation metric with further details provided in the Appendix.

\paratitle{Backbone Models.}
Experiments are conducted on three widely used RoPE-based LLM families with different architectures and design choices, namely \textbf{LLaMA~3.2}, \textbf{Qwen3}, and \textbf{Gemma~3}~\cite{grattafiori2024llama,qwen3technicalreport,gemma_2025}, to evaluate the robustness and generality of DyPAM.

\paratitle{Baseline Methods.}
We compare DyPAM with a diverse set of PEFT methods that adopt different adaptation strategies.
Low-rank adaptation methods include LoRA~\cite{hu2021lora} and AdaLoRA~\cite{zhang2023adalora}, which parameterize weight updates in a low-dimensional subspace, with AdaLoRA further adjusting rank allocation across layers according to their relative importance.
We also consider structured weight reparameterization approaches, including OFT~\cite{qiu2023controlling} and Bone~\cite{kang2024balancing}, where OFT constrains updates to orthogonal transformations and Bone employs block-wise affine parameterization to capture structured correlations within weight matrices.
In addition, we include lightweight modulation-based methods such as IA$^3$~\cite{liu2022few} and LNTuning~\cite{zhao2023tuning}, which adapt the model by rescaling internal activations or normalization parameters with minimal trainable parameters.
We further compare against FourierFT~\cite{gao2024parameter}, which performs adaptation in the frequency domain by learning a compact set of spectral coefficients for weight updates.
Finally, we include SHiRA~\cite{shiracite}, which applies sparse high-rank adapters to update only a small subset of backbone weights, and RoSA~\cite{pan2026rosa}, which performs RoPE-aware selective adaptation over attention dimensions and layers.

\paratitle{Implementation Details.}
All experiments are conducted using DeepSpeed~\cite{rasley2020deepspeed} with bfloat16 precision on NVIDIA RTX 4090 GPUs.
For DyPAM, we use a modulation embedding dimension $d_e = 64$, a low-rank projection rank $r = 128$, and a modulation strength $\alpha = 0.3$.
For baseline methods, except for extremely low-parameter approaches, we match the number of trainable parameters to a comparable scale. 
Additional implementation details are provided in the appendix, and all code and data are released to ease reproducibility\footnote{\codelink}.

\begin{table}[t]
    \centering
    \caption{Macro-averaged accuracy on mathematical reasoning benchmarks across Qwen3 model scales. The comparison includes representative strong PEFT baselines from the main experiments.}
    \resizebox{1\linewidth}{!}{
        \renewcommand{\arraystretch}{1}
        \begin{tabular}{lcccc}
            \toprule
            \textbf{Baseline} 
            & \textbf{Qwen 3 0.6B} 
            & \textbf{Qwen 3 1.7B} 
            & \textbf{Qwen 3 4B}
            & \textbf{Qwen 3 8B} \\
            \midrule
            LoRA            
                & 64.06
                & 66.64
                & 75.60 
                & 80.37 \\
            OFT         
                & \underline{65.96}
                & \underline{67.81}
                & 75.54
                & 80.45 \\
            SHiRA         
                & 63.95
                & 64.65 
                & 73.33
                & 81.04 \\
            RoSA         
                & 63.99
                & 67.38 
                & \underline{77.92} 
                & \underline{81.29} \\
            \framework (ours) 
                & \textbf{66.13} 
                & \textbf{69.24} 
                & \textbf{78.24} 
                & \textbf{83.20}  \\
            \bottomrule
        \end{tabular}
    }
    \label{tab:scale}
    \vspace{-9px}
\end{table}

\begin{table*}[t]
    \centering
    \small
    \caption{Comparison of \framework with PEFT baselines on commonsense reasoning benchmarks across three backbone models.
Micro-avg and macro-avg denote micro- and macro-averaged performance.
Best results are highlighted in bold, and second-best results are underlined.
$^{*}$ indicates statistically significant improvements over the best baseline (two-sided t-test, $p<0.05$).}
    \resizebox{1\linewidth}{!}{
        \renewcommand{\arraystretch}{0.97}
        \begin{tabular}{l|l ccccccccc cc}
                \toprule
            \textbf{Backbone LLM} & \textbf{Method} 
            & \textbf{Param(\%)}
            & \textbf{BoolQ} & \textbf{PIQA} & \textbf{SocialIQA}
            & \textbf{ARC-C} & \textbf{ARC-E} & \textbf{OpenBookQA}
            & \textbf{HellaSwag} & \textbf{WinoGrande}
            & \textbf{Micro Avg} & \textbf{Macro Avg} \\
            \midrule
            \multirow{10}{*}{\textbf{LLaMA 3.2 3B}}
                & LoRA      & 1.12 &63.61&\underline{79.71}&66.94&69.45&84.05&67.00&73.94&55.56&71.94&70.03 \\
                & AdaLoRA   & 2.22 &63.52&78.94&67.09&68.94&\underline{85.14}&70.20&78.11&56.35&73.95&71.04 \\
                & OFT       & 0.73 &\underline{65.63}&79.54&\underline{70.37}&\underline{70.39}&85.06&\textbf{71.80}&83.15&\textbf{66.38}&\underline{77.52}&\underline{74.04} \\
                & Bone      & 1.14 &64.56&75.68&69.34&64.42&79.76&70.20&75.92&\underline{65.75}&72.77&70.70 \\
                & IA$^3$    & 0.02 &62.32&77.09&59.67&57.94&77.10&57.40&50.48&52.25&58.66&61.78 \\
                & LN Tuning & 0.01 &62.51&76.99&59.52&59.81&76.52&59.00&52.02&52.17&59.42&62.32 \\
                & FourierFT & 0.73 &62.14&79.49&61.98&61.86&80.93&62.40&73.21&49.09&69.75&66.39 \\
                & SHiRA     & 1.12 &65.23&79.65&69.14&\textbf{71.16}&84.97&71.20&\underline{83.18}&65.67&77.35&73.78 \\
                & RoSA      & 0.54 &64.53&79.65&69.86&69.28&84.43&70.80&83.12&63.54&77.00&73.15 \\
                & \framework (ours)     & 0.92 &\textbf{65.93}&\textbf{79.76}&\textbf{70.88}&\underline{70.39}&\textbf{85.19}
                              &\textbf{71.80}&\textbf{83.71}&65.35&\textbf{77.83*}&\textbf{74.13*} \\
            \midrule
            \multirow{10}{*}{\textbf{Qwen3 8B}}
                & LoRA      & 0.79 &70.49&86.34&77.18&90.19&96.51&87.60&89.50&72.85&85.19&83.83 \\
                & AdaLoRA   & 1.57 &70.73&86.51&76.71&\underline{90.36}&96.55&87.20&88.92&72.38&84.91&83.67 \\
                & OFT       & 0.51 &69.97&86.83&76.56&89.93&\underline{96.97}&88.00&89.17&76.48&85.20&84.24 \\
                & Bone      & 0.81 &69.02&85.31&75.64&88.91&95.58&87.60&89.30&\underline{76.56}&84.71&83.49 \\
                & IA$^3$    & 0.02 &69.51&86.34&76.71&90.27&96.09&84.40&85.12&66.77&82.59&81.90 \\
                & LN Tuning & 0.00 &69.33&86.40&75.95&90.27&96.00&83.00&83.86&65.43&81.82&81.28 \\
                & FourierFT & 0.37 &69.54&84.49&73.13&85.92&95.29&77.80&80.48&62.27&79.34&78.62 \\
                & SHiRA     & 0.79 &\underline{70.83}&\underline{87.05}&\textbf{77.33}&\underline{90.36}&\underline{96.97}&\underline{88.20}
                              &\textbf{89.56}&75.77&\underline{85.57}&\underline{84.51} \\
                & RoSA      & 0.36 &68.96&86.94&75.33&89.85&96.38&\underline{88.20}&89.43&76.16&84.99&83.91 \\
                & \framework (ours)     & 0.61 &\textbf{70.89}&\textbf{87.11}&\textbf{77.33}&\textbf{90.53}&\textbf{97.05}
                              &\textbf{88.80}&\underline{89.53}&\textbf{76.80}&\textbf{85.66*}&\textbf{84.75*} \\
            \midrule
            \multirow{10}{*}{\textbf{Gemma 3 4B}}
                & LoRA      & 1.33 &65.72&79.71&69.40&74.49&87.08&71.00&74.53&55.01&73.37&72.12 \\
                & AdaLoRA   & 2.62 &\underline{66.09}&79.49&68.73&76.54&89.02&74.00&73.20&58.09&73.30&73.14 \\
                & OFT       & 0.75 &65.69&81.99&74.51&\underline{76.71}&88.47&78.00&\underline{83.86}&\underline{65.27}&\underline{79.17}&\underline{76.81} \\
                & Bone      & 1.41 &64.68&75.35&71.24&70.39&82.83&75.80&78.33&64.48&74.70&72.89 \\
                & IA$^3$    & 0.02 &62.17&71.49&57.32&57.51&73.19&55.20&44.89&57.85&55.30&59.95 \\
                & LN Tuning & 0.00 &62.60&66.70&49.85&49.91&63.59&45.20&47.29&60.46&53.90&55.70 \\
                & FourierFT & 0.37 &63.94&75.57&67.14&67.32&76.05&57.80&71.81&59.35&69.76&67.37 \\
                & SHiRA     & 0.79 &65.57&\underline{82.25}&\underline{74.53}&76.19&\textbf{89.71}&\underline{78.20}&83.19&64.48&78.94&76.77 \\
                & RoSA      & 0.40 &63.70&79.54&67.40&72.27&86.66&69.40&48.53&47.51&60.62&66.88 \\
                & \framework (ours)     & 0.62 &\textbf{66.21}&\textbf{82.59}&\textbf{74.82}&\textbf{77.13}&\underline{89.23}
                              &\textbf{79.20}&\textbf{84.09}&\textbf{65.35}&\textbf{79.56*}&\textbf{77.33*} \\
            \bottomrule
        \end{tabular}
    }
    \label{tab:main_common}
\end{table*}

\subsection{Overall Performance (RQ1)}

Across both mathematical and commonsense reasoning benchmarks, \framework consistently outperforms existing PEFT baselines under comparable parameters, providing a clear positive answer to \textbf{RQ1}. The improvements are observed across all evaluated backbone models and task categories, indicating that the effectiveness of \framework generalizes across different models and datasets.

Specifically, on mathematical reasoning benchmarks as shown in Table~\ref{tab:main_math}, \framework demonstrates strong and stable gains across heterogeneous tasks that require multi-step computation, and numerical reasoning.
The improvements are reflected in both micro- and macro-averaged metrics, suggesting that \framework enhances overall reasoning capability rather than favoring a small subset of benchmarks.
In contrast, existing PEFT baselines exhibit more uneven behavior. Low-rank methods such as LoRA and AdaLoRA tend to yield limited gains, particularly on more challenging datasets.
Frequency-domain approaches such as FourierFT show moderate performance but lack robustness across datasets. Methods with extremely small parameter budgets, including IA$^3$ and LN-Tuning, generally underperform due to their limited adaptation capacity. 
Structured or orthogonal methods such as OFT and Bone often perform well on specific tasks but struggle to maintain consistent improvements across the full benchmark suite, whereas sparse adaptation methods like SHiRA, which selectively activate a subset of parameters during fine-tuning, achieve strong performance on multiple tasks, highlighting the potential of structured parameter updates. RoSA also benefits from structured adaptation by introducing rotational subspace updates, leading to competitive results on several benchmarks.
However, RoSA applies a static RoPE-aware selection that is shared across all inputs, whereas \framework modulates query and key dimension pairs in an input-conditioned, token-dependent manner coordinated with head-wise and layer-wise structure.
This fine-grained and input-conditioned modulation, rather than the use of positional structure alone, accounts for the consistent margin of \framework over RoSA across backbones.
Compared with these baselines, \framework achieves more balanced improvements across tasks and models, indicating superior robustness and generalization.

A similar pattern is observed on commonsense reasoning tasks as shown in Table~\ref{tab:main_common}. \framework yields balanced improvements across diverse datasets spanning factual verification and social commonsense. The consistent gains in macro-averaged performance indicate improved robustness at the task level, while the improvements in micro-averaged accuracy reflect stronger overall performance.

Overall, these results show that \framework provides a more reliable and general-purpose adaptation mechanism than prior PEFT approaches. By delivering consistent gains across tasks, domains, and backbone models, \framework effectively addresses \textbf{RQ1} and demonstrates clear advantages over existing PEFT baselines.

\subsection{Scalability Analysis (RQ2)}

Table~\ref{tab:scale} evaluates the scalability of \framework across Qwen3 models from 0.6B to 8B using macro-averaged accuracy on mathematical reasoning benchmarks. As model size increases, \framework consistently achieves better performance than the strongest PEFT baselines at each scale.
Moreover, the performance difference between \framework and baselines becomes larger as the backbone grows, indicating that \framework continues to benefit from increased capacity.

Overall, these results demonstrate that \framework maintains strong scalability across model sizes and remains effective when applied to both small and large backbone models, directly addressing \textbf{RQ2}.

\subsection{Ablation and Sensitivity Analysis (RQ3, 4)}
We further analyze \framework through ablation studies and parameter sensitivity experiments, aiming to understand the contribution of individual components and the robustness of key hyperparameters.

The ablation results show that each core component of \framework plays a complementary role in overall performance. Removing any single component consistently leads to performance degradation, indicating that the gains of \framework arise from their joint design rather than isolated architectural choices.
In addition, the sensitivity analysis on the modulation strength $\alpha$ demonstrates the effectiveness of this design choice. Proper modulation significantly improves performance compared to weak or overly strong modulation.
Overall, these analyses confirm that the performance improvements of \framework are structurally grounded and robust.

\subsection{Analysis of Modulation Patterns (RQ5)}

We analyze the learned modulation from two perspectives: the structural bias terms and the effective modulation ranges.

\begin{figure}[t]
    \centering
    \includegraphics[width=1\linewidth]{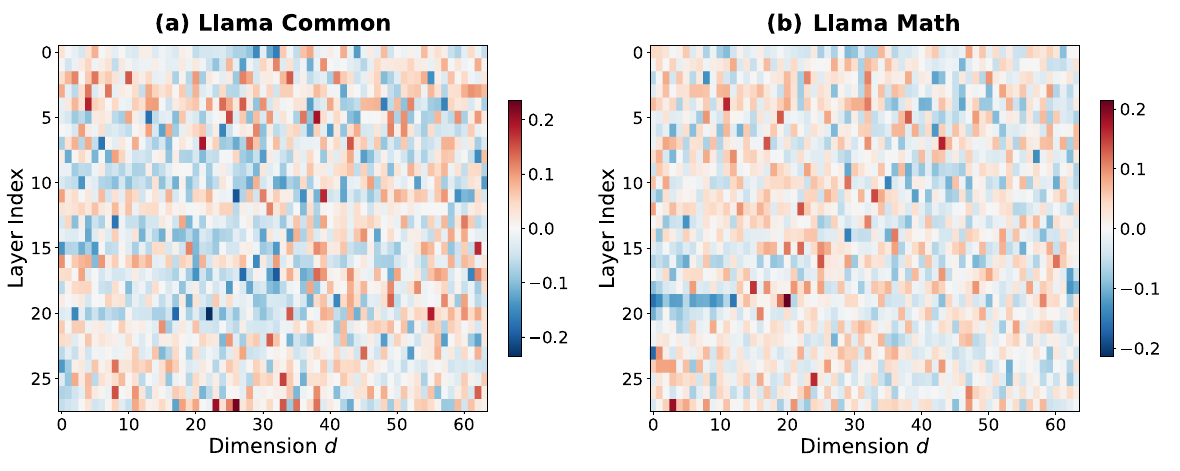}
    \caption{Learned layer-wise query bias over attention dimensions on LLaMA-3.2-3B under commonsense and mathematical reasoning settings.
The heatmaps show that the structural bias varies across layers and dimensions rather than following a uniform pattern.}
    \label{fig:bias_pattern}
  \vspace{-2px}
\end{figure}

Figure~\ref{fig:bias_pattern} visualizes the learned layer-wise query bias over attention dimensions on LLaMA-3.2-3B under commonsense and mathematical reasoning settings.
Rather than exhibiting uniform shifts, the bias values vary across both layers and dimensions, indicating that \framework learns heterogeneous, dimension-specific adjustments.
This structured non-uniformity suggests that different attention dimensions develop distinct positional preferences at different depths, aligning with the intuition that positional information is utilized differently across layers.
In addition, faint horizontal band patterns can be observed, hinting that certain layers may exhibit 
consistent preferences over specific subsets of attention dimensions.

\begin{figure}[t]
    \centering
    \includegraphics[width=1\linewidth]{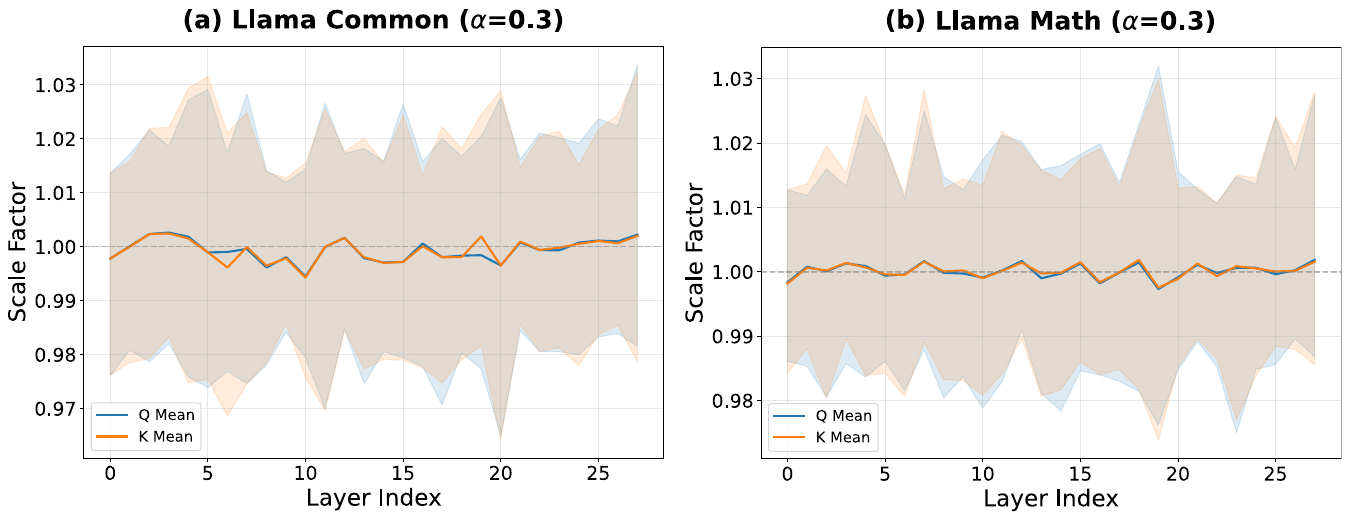}
    \caption{Layer-wise modulation range of query and key representations on LLaMA-3.2-3B under commonsense and mathematical reasoning settings.
The mean scale remains close to 1, while the effective range changes across layers and training data.}
    \label{fig:modulation_range}
% \vspace{-4px}
\end{figure}

Figure~\ref{fig:modulation_range} summarizes the effective modulation range of query and key representations under the same two settings.
The scale factors remain centered around 1, while the range varies across layers and training data.
This restrained behavior preserves the pretrained attention structure while allowing flexibility.

Overall, these visualizations demonstrate that \framework learns structured and stable positional modulation patterns.
By combining dimension-wise modulation with restrained head-wise and layer-wise scaling, \framework adapts positional attention in a targeted manner, shaping how token dependencies are formed and thereby helping explain its consistent performance gains across models and tasks.

\begin{figure}[t]
    \centering
    \includegraphics[width=1\linewidth]{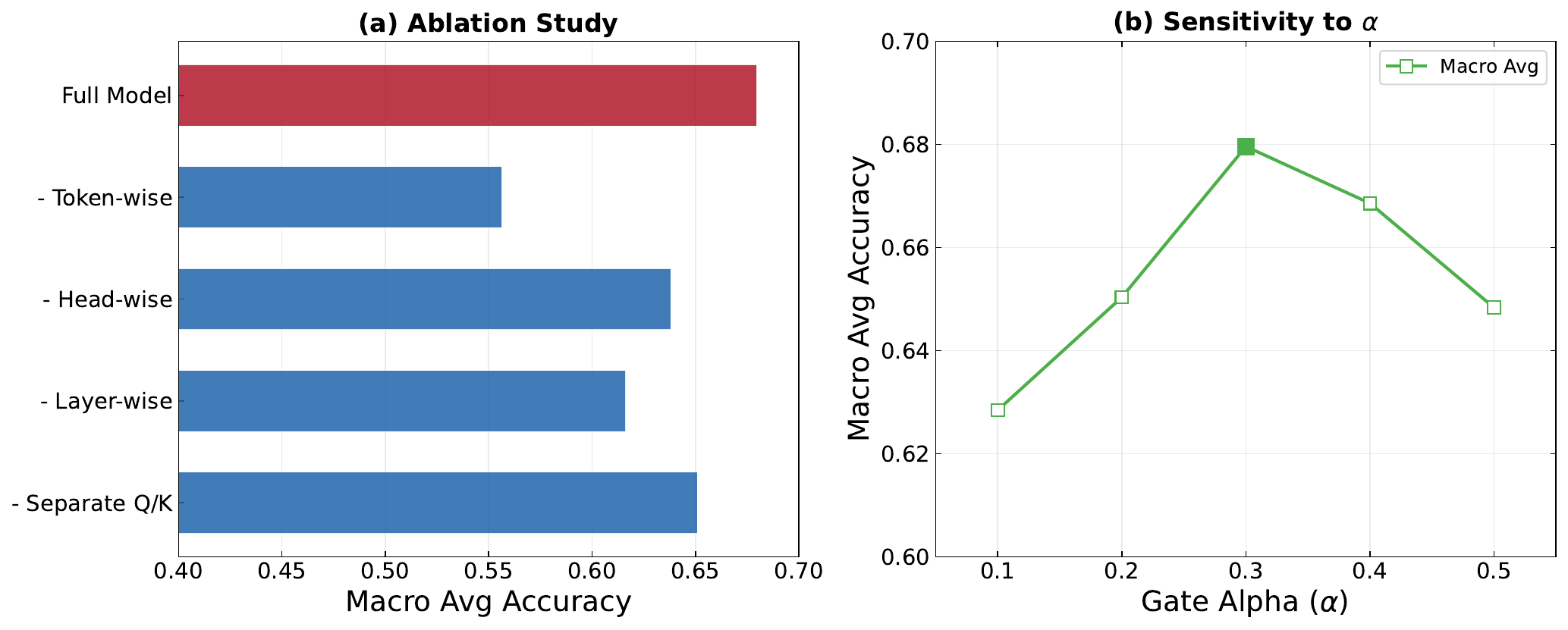}
    \caption{Ablation and hyperparameter sensitivity of DyPAM.
The ablation removes individual modulation components, while the sensitivity study varies the modulation strength $\alpha$.}
    \label{fig:ablation_sensitivity}
  \vspace{-5px}
\end{figure}

\section{Related Work}

\subsection{Parameter-Efficient Fine-Tuning}
Recent advancements in PEFT focus on reducing the number of trainable parameters while maintaining performance. Low-rank approaches such as LoRA~\cite{hu2021lora} and AdaLoRA~\cite{zhang2023adalora} adapt pre-trained weights through additive low-rank updates, while scaling-based methods including IA$^3$~\cite{liu2022few} and LNTuning~\cite{zhao2023tuning} employ lightweight gating to modulate activations under strict parameter constraints. Structured PEFT approaches, such as OFT~\cite{qiu2023controlling} and Bone~\cite{kang2024balancing}, further constrain the update space by enforcing geometric properties, whereas spectral-domain methods like FourierFT~\cite{gao2024parameter} perform adaptation via frequency-based transformations of model representations. In addition, RoSA~\cite{pan2026rosa} is motivated by the structure of RoPE and proposes RoPE-aware selective adaptation by emphasizing low-frequency attention components. Compared with RoSA, which applies static RoPE-aware selection, \framework generates input-conditioned modulation over query and key dimension pairs. In contrast to these approaches, our method explicitly models fine-grained, input-conditioned modulation within attention, enabling structure-aware adaptation.

\subsection{Structured Modeling in Transformer}
Transformer attention has been recognized for its structural heterogeneity, with different attention heads specializing in various semantic or syntactic aspects of the input~\cite{raganato2020fixed}.
Recent works have analyzed head-level specialization and layer-wise differences, demonstrating that attention heads should not be treated uniformly~\cite{voita2019analyzing,zhang2022mixture}.
Beyond language modeling, attention- and frequency-aware designs have also been widely adopted in spatiotemporal modeling~\cite{jiang2023pdformer,ma2026hierarchical} and financial prediction~\cite{wang2019alphastock,wang2021impact}, further underscoring the value of structure-aware mechanisms.
Positional encodings, particularly in RoPE and its variants, are not just external biases but integral to the attention computation, providing a way to model relative positional relationships more effectively~\cite{gu2025unpacking}.
Newer methods, such as ComRoPE~\cite{yu2025comrope}, introduce dynamic and input-conditioned positional encodings, enhancing the flexibility of position representations in transformers.
RoSA~\cite{pan2026rosa} is also motivated by the structure of RoPE and explores selective adaptation of attention components guided by frequency characteristics.
While these studies highlight the structured and positional nature of attention, they largely rely on static or coarse-grained mechanisms, motivating the need for fine-grained and input-conditioned modulation within attention.
 
\section{Conclusion}

In this work, we introduce DyPAM, a PEFT method that adapts LLMs by modulating positional attention in a fine-grained and structured manner.
Motivated by heterogeneous attention behavior across dimensions, heads, layers, and input tokens, DyPAM operates directly on the query and key representations and aligns adaptation with the RoPE-induced positional structure.
It jointly models input-conditioned, dimension-wise and head-wise, layer-wise structural modulation, providing a principled approach to adapting positional attention within the PEFT paradigm.
By reshaping how positional information influences attention, DyPAM adjusts the way dependencies are formed across tokens, which is consistent with its gains on tasks that rely on multi-step and relational reasoning.
Extensive experiments across mathematical and commonsense reasoning benchmarks demonstrate that DyPAM consistently outperforms strong PEFT baselines.
Future work will further optimize the runtime implementation and extend DyPAM across diverse architectural mechanisms.

\begin{acks}
This work is supported by the National Natural Science Foundation of China (No. 725B2004, 72242101, 72222022), and the Science and Technology Development Fund Macau SAR (0052/2023/RIA1).
\end{acks}

\clearpage
\bibliographystyle{ACM-Reference-Format}
\bibliography{ref}

\appendix
\section{Attention Activation Patterns}
\label{sec:appendix_activation}

\subsection{Effect of Positional Encoding Schemes}
\label{sec:appendix_posenc}
\begin{figure}[t]
    \centering
    \includegraphics[width=1\linewidth]{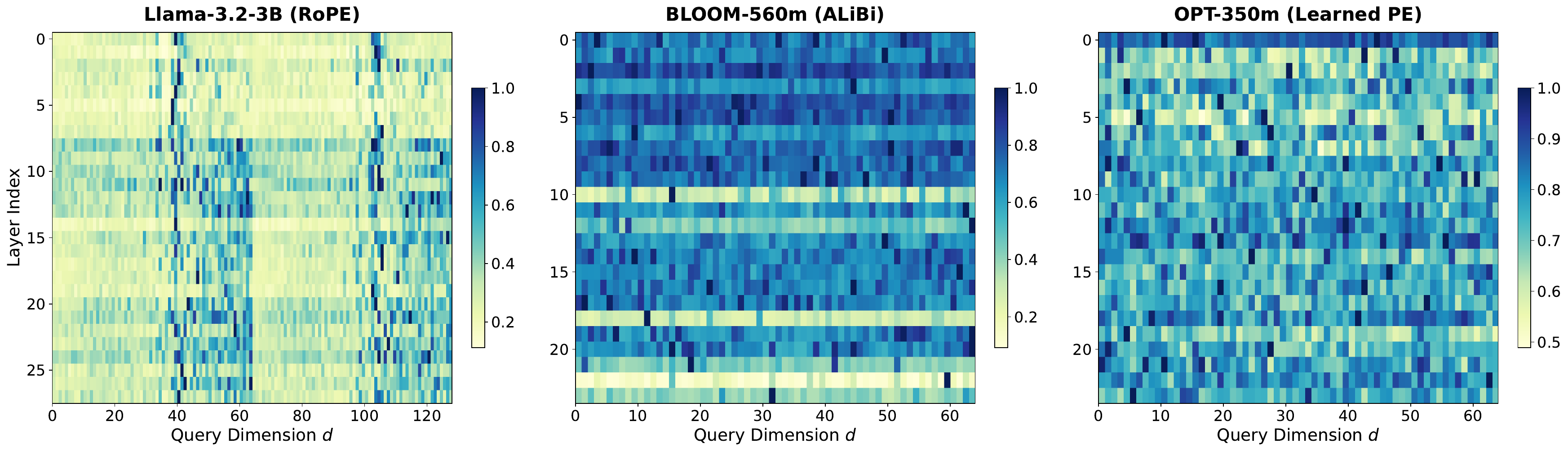}
    \caption{Cross-model comparison of query activation patterns across layers and attention dimensions. RoPE-based LLaMA-3.2-3B shows structured, dimension-dependent patterns, while ALiBi-based BLOOM-560M and OPT-350M with learned embeddings are more homogeneous.}
    \label{fig:appendix_posenc_compare}
\end{figure}

To verify that the heterogeneous attention behavior motivating \framework is tied to the positional encoding mechanism, we compare query activation patterns across models using different positional encodings, including RoPE, ALiBi, and learned positional embeddings.
Specifically, we visualize query representations from LLaMA-3.2-3B, BLOOM-560M, and OPT-350M under a unified scheme covering all attention dimensions (Figure~\ref{fig:appendix_posenc_compare}).
RoPE-based models exhibit structured, dimension-dependent activation patterns, whereas models using ALiBi or learned positional embeddings display substantially more homogeneous behavior across dimensions, supporting our focus on RoPE-induced positional structure.

\subsection{Token-Dependent Activation Patterns}
\label{sec:appendix_tokenwise_sub}
\begin{figure}[t]
    \centering
    \includegraphics[width=1\linewidth]{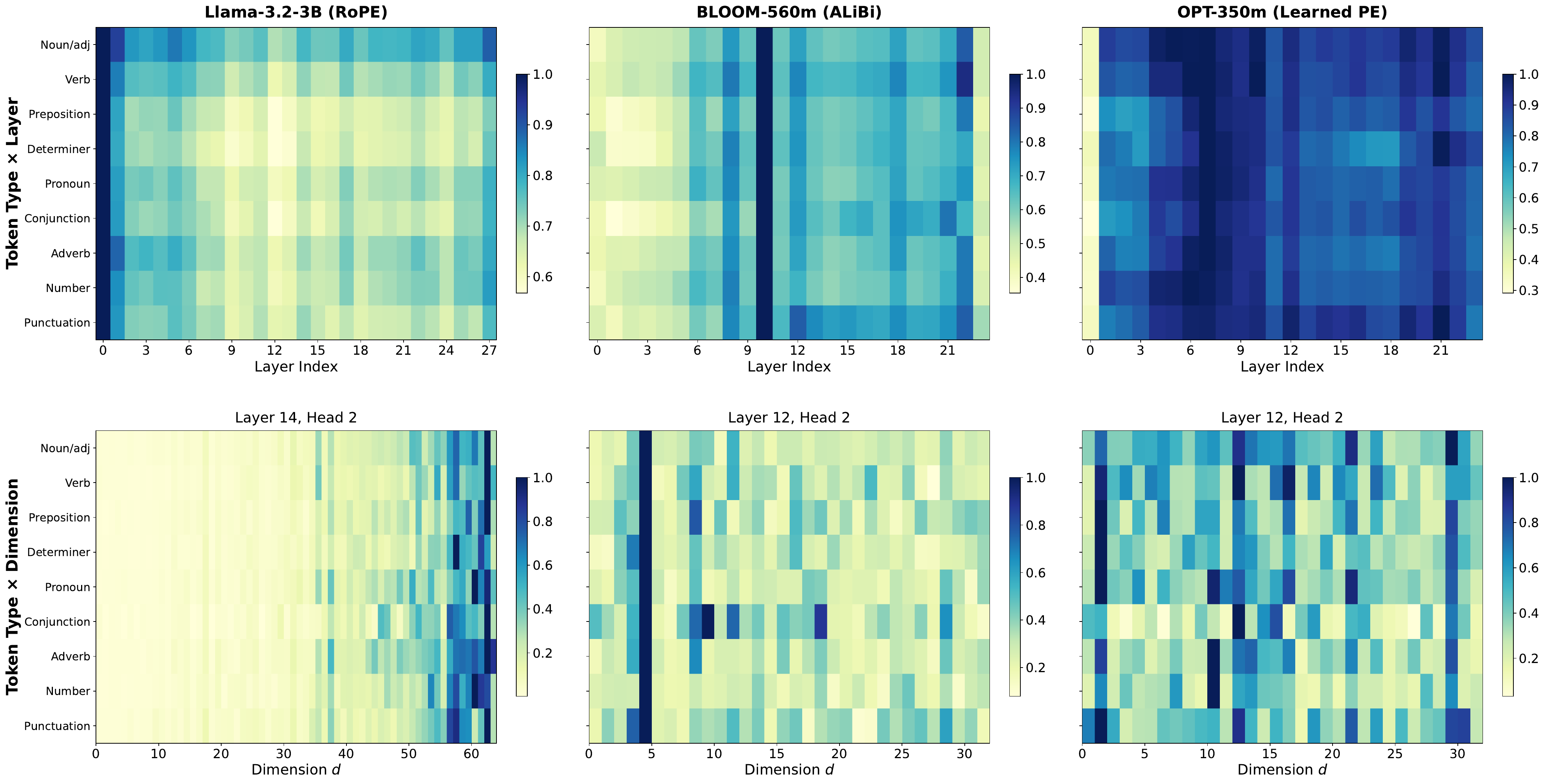}
    \caption{Token-type activation patterns across layers and attention dimensions for LLaMA-3.2-3B, BLOOM-560M, and OPT-350M. Token-dependent variation is most pronounced in the RoPE-based model.}
    \label{fig:appendix_tokenwise}
\end{figure}

We further examine token-dependent variation in attention activations.
We visualize query activations conditioned on different token types and input contexts, extending the analysis of Fig.~\ref{fig:rope_channel}(c,d) to multiple models (Figure~\ref{fig:appendix_tokenwise}).
In the RoPE-based model, different tokens induce systematically different activation patterns, while such token-dependent structure is less pronounced under alternative positional encoding schemes.
These observations motivate the input-conditioned, token-level component of \framework.

\section{Experimental and Implementation Details}
\label{sec:appendix_impl}
All experiments reported in the main paper use a primary random seed of 2333, with additional runs using seeds 1000, 2000, 3000, and 4000 to assess statistical significance and reproducibility.
Training employs mixed-precision (BF16) and DeepSpeed~\cite{rasley2020deepspeed} with ZeRO Stage~1.
A randomly selected validation set of 300 samples from the training data is used for checkpoint selection, and the checkpoint with the lowest validation loss is used for testing.

The default hyperparameter settings are summarized below:
\begin{itemize}[leftmargin=*, topsep=0pt]
\item Optimizer: AdamW; learning rate $1\times10^{-3}$ with cosine schedule; warmup ratio 0.05; weight decay 0.
\item Batch size: 2 (gradient accumulation steps of 2); max sequence length: 2048.
\item Modulation feature dimension $d_e=64$; low-rank projection rank $r=128$; modulation strength $\alpha=0.3$.
\end{itemize}

\subsection{Software and Environment}
The software environment uses PyTorch 2.7.0, DeepSpeed 0.18.4, NumPy 2.2.6, PEFT 0.18.1, Transformers 4.51.0, Tokenizers 0.21.4, and CUDA 12.8.
Experiments run on Ubuntu 24.04 LTS with an Intel Xeon Gold 6330 CPU, an NVIDIA GeForce RTX 4090 GPU, and 512GB RAM.
For baseline methods, except for extremely low-parameter approaches, we match the number of trainable parameters to a comparable scale; all other hyperparameters follow their default configurations from the PEFT library~\cite{peft} or the original implementations.
Detailed implementation and datasets can be found in our codebase\footnote{\codelink}.

\section{Dataset Details}
\label{sec:appendix_dataset}
We use two instruction-tuning datasets from the LLM-Adapters benchmark suite~\cite{hu2023llm}: \textbf{Math10K} for math word problems with step-by-step solutions and \textbf{Commonsense15K} for commonsense reasoning questions normalized into a consistent instruction format. Statistics are summarized in Table~\ref{tab:dataset}.

\begin{table}[t]
    \centering
	\small
	\caption{Statistics of the training datasets for commonsense and mathematical reasoning tasks.}
    \resizebox{0.95\linewidth}{!}{
    \renewcommand{\arraystretch}{1.01}
    \begin{tabular}{lccc}
        \toprule
        \textbf{Dataset}     & \textbf{Samples}     & \textbf{Total Tokens} & \textbf{Avg. Tokens/Sample}  \\
        \midrule
        Math10K        & 9,919          &   2,273,016                & 229.16 \\
        Commonsense15K        & 15,119          &   1,778,782                & 117.65 \\
        \bottomrule
    \end{tabular}
    }
    \label{tab:dataset}
\end{table}

We evaluate on well-established commonsense and mathematical reasoning benchmarks; detailed statistics are in Table~\ref{tab:arith-datasets}~(Mathematical) and Table~\ref{tab:commonsense-datasets}~(Commonsense).

\noindent \textbf{a) Mathematical Reasoning:} MultiArith~\cite{roy2016solving} (multi-step arithmetic), GSM8K~\cite{cobbe2021training} (grade-school multi-step reasoning), AddSub~\cite{hosseini2014learning} (addition/subtraction), AQuA~\cite{ling2017program} (algebraic problems with rationales), SingleEq~\cite{koncel2015parsing} (equation-tree parsing), SVAMP~\cite{patel2021nlp} (structural variations), and MAWPS~\cite{koncel2016mawps} (unified word-problem benchmark).

\begin{table}[t]
\centering
\small
\caption{Statistics of Mathematical Reasoning Test Datasets.}
\resizebox{0.8\linewidth}{!}{
    \renewcommand{\arraystretch}{0.95}
	\begin{tabular}{lcc}
	\toprule
	\textbf{Dataset} & \textbf{Samples} & \textbf{Answer Type} \\
	\midrule
	MultiArith & 600   & Numeric \\
	GSM8K      & 1,319 & Numeric \\
	AddSub     & 395   & Numeric \\
	AQuA       & 254   & Multiple Choice (A--E) \\
	SingleEq   & 508   & Numeric \\
	SVAMP      & 1,000 & Numeric \\
	MAWPS      & 238   & Numeric \\
	\bottomrule
	\end{tabular}
}
\label{tab:arith-datasets}
\end{table}

\noindent \textbf{b) Commonsense Reasoning:} BoolQ~\cite{clark2019boolq} (yes/no QA), PIQA~\cite{bisk2020piqa} (physical commonsense), SIQA~\cite{sap2019socialiqa} (social commonsense), ARC-Challenge/ARC-Easy~\cite{clark2018think} (science QA), OBQA~\cite{mihaylov2018can} (open-book science QA), HellaSwag~\cite{zellers2019hellaswag} (adversarial NLI), and WinoGrande~\cite{sakaguchi2020winogrande} (pronoun resolution).

\begin{table}[t]
\centering
\small
\caption{Statistics of Commonsense Reasoning Test Datasets.}
\resizebox{1\linewidth}{!}{
    \renewcommand{\arraystretch}{1.01}
	\begin{tabular}{lcc}
	\toprule
	\textbf{Dataset} & \textbf{Samples} & \textbf{Answer Format} \\
	\midrule
	BoolQ         & 3,270  & true / false \\
	PIQA          & 1,838  & solution1 / solution2 \\
	SIQA    & 1,954  & answer1 / answer2 / answer3 \\
	ARC-Challenge & 1,172  & answer1 / answer2 / answer3 / answer4 \\
	ARC-Easy      & 2,376  & answer1 / answer2 / answer3 / answer4 \\
	OBQA    & 500    & answer1 / answer2 / answer3 / answer4 \\
	HellaSwag     & 10,042 & ending1 / ending2 / ending3 / ending4 \\
	WinoGrande    & 1,267  & option1 / option2 \\
	\bottomrule
	\end{tabular}
}
\label{tab:commonsense-datasets}
\end{table}

\section{Evaluation Protocol and Metrics}
\label{sec:appendix_eval}
All model outputs are generated with greedy decoding~(\texttt{do\_sample}= False, \texttt{max\_new\_tokens=256}) via the \texttt{generate()} API in Hugging Face Transformers, using the unified instruction template below.
\begin{tcolorbox}[boxrule=0.8pt]
\textless s\textgreater Below is an instruction that describes a task. Write a response that appropriately completes the request.

\#\#\# Instruction:\\
\{instruction\}
\\
\\
\#\#\# Response:
\end{tcolorbox}
Predictions are extracted with task-specific regular expressions: for mathematical reasoning, numerical answers (absolute tolerance $10^{-3}$) or alphabetic choices (A--E) for AQuA; for commonsense reasoning, exact-match answers (true/false and option labels) against ground-truth labels. All extraction and accuracy computation scripts are provided in our codebase.

\begin{table}[t]
    \centering
    \small
    \caption{Inference overhead of \framework compared with the base model and representative PEFT baselines on LLaMA-3.2-3B. Latency is reported per token. The per-token value is lower at batch size $32$ because more tokens are processed in parallel.}
    \resizebox{1\linewidth}{!}{
    \renewcommand{\arraystretch}{1.05}
    \begin{tabular}{lcccc}
        \toprule
        \textbf{Method} & \textbf{BS=1 (ms/tok)} & \textbf{BS=32 (ms/tok)} & \textbf{Mem (GB)} & \textbf{FLOPs overhead} \\
        \midrule
        Base            & 14.49 & 0.49 & 6.00 & --- \\
        LoRA (unmerged) & 27.64 & 1.52 & 6.06 & 73.0M (+1.13\%) \\
        RoSA            & 22.40 & 1.45 & 6.03 & 34.9M (+0.54\%) \\
        \framework (ours) & 24.98 & 1.49 & 6.06 & 66.1M (+1.03\%) \\
        \bottomrule
    \end{tabular}
    }
    \label{tab:inference_overhead}
\end{table}

\section{Inference Overhead}
\label{sec:appendix_overhead}
We report the inference overhead of \framework against the base model and representative PEFT baselines in Table~\ref{tab:inference_overhead}, measured on LLaMA-3.2-3B.
\framework adds $1.03\%$ FLOPs overhead, lower than unmerged LoRA ($1.13\%$), as it avoids low-rank updates to all linear layers and the per-token modulation introduces only a small cost.
\framework is about $10\%$ faster than unmerged LoRA at batch size $1$ and comparable at batch size $32$, with negligible memory overhead.
Unlike LoRA, \framework cannot be merged into the backbone weights due to its input-conditioned design, but the overhead remains controlled and can be further reduced with optimizations such as KV-cache reuse and kernel fusion.

          % slim appendix, follows references
% \clearpage
% \clearpage
% \input{sections/99_appendix_archive.tex}  % full archived appendix, preview only
\end{document}